\documentclass[conference]{IEEEtran}
\IEEEoverridecommandlockouts
\usepackage{cite}
\usepackage{amsmath,amssymb,amsfonts}
\usepackage{algorithmic}
\usepackage{graphicx}
\usepackage{textcomp}
\usepackage{xcolor}
\def\BibTeX{{\rm B\kern-.05em{\sc i\kern-.025em b}\kern-.08em
    T\kern-.1667em\lower.7ex\hbox{E}\kern-.125emX}}
\usepackage[bookmarks=false]{hyperref}
\usepackage{url}
\usepackage{times}
\usepackage{helvet}
\usepackage{courier}
\usepackage{mathtools}
\usepackage{amsfonts}       
\usepackage{subcaption}
\usepackage{booktabs}       
\usepackage{tikz}
\usepackage{wrapfig,lipsum,booktabs}
\usepackage{physics}
\begin{document}

\title{The Variational InfoMax AutoEncoder}


\author{Vincenzo Crescimanna, Bruce Graham \\
Department of Computer Science\\
University of Stirling\\
Stirling, UK \\
\texttt{\{vincenzo.crescimanna1, bruce.graham\}@stir.ac.uk}}

%


\maketitle

\begin{abstract}
The Variational AutoEncoder (VAE) learns simultaneously an inference and a generative model, but only one of these models can be learned at optimum, this behaviour is associated to the ELBO learning objective, that is optimised by a non-informative generator. In order to solve such an issue, we provide a learning objective, learning a maximal informative generator while maintaining bounded the network capacity: the Variational InfoMax (VIM).
The contribution of the VIM derivation is twofold: an objective learning both an optimal inference and generative model and the explicit definition of the network capacity, an estimation of the network robustness.

\end{abstract}

\section{Introduction}

A common assumption in machine learning is that any visible data $x \in \mathcal{X}$ is completely described by some generative factor $o$, living in a smaller hidden space $\mathcal{O}$, i.e. $x = g(o)$ with $g$ a (possibly stochastic) generative function. The aim of unsupervised representation learning research is to find a \emph{representation} $z$ of the generative factor $o$ living in a known space $\mathcal{Z}$ describing, as well as $o$, the visible data $x$. The relevance of such task is twofold: since the learnt small representation $z$ is task agnostic it can be used as input for networks performing different tasks (\emph{generalisation property}), \cite{rifai2011higher}, and also because such representation allows to interpret what is learning the network in its hidden layers \cite{lipton2018mythos}.

Many models $f_\phi : \mathcal{X} \to \mathcal{Z}$, parametrising an inference distribution $q_\phi(z|x)$, have been proposed \cite{dinh2016density,hinton2006fast,maddison2017filtering,radford2015unsupervised}, but recently in order to solve this problem it was proposed to consider a dual problem: define a priori  $z$  and find a generator map $g_\theta$, such that for any $z$, $g_\theta(z)$ is an element of $\mathcal{X}$. In particular, two families of probabilistic generative models have become dominant: Variational AutoEncoder (VAE) \cite{kingma2013auto,rezende2014stochastic} and Generative Adversarial Network (GAN) \cite{goodfellow2014generative}. 
The common idea of the two approaches is that a good generator $p_\theta(x|z)$ is the one able to generate the data that is as close as possible to the visible one, i.e. that with respect a certain metric $D$, the distance between the marginal $p_\theta(x) = \mathbb{E}_{p(z)}[p_\theta(x|z)]$ and the visible distribution $p_{D}(x)$ is minimal.

In this manuscript we restrict our attention to the VAE model, since by its architecture, it is the only one learning an inference model $q_\phi(z|x)$, and where the learnt representation, possibly from different datasets, can be used as input for networks performing different tasks, \cite{achille2018life,ramapuram2017lifelong}. Although VAE, by its  training robustness and general good generative performance is the most popular representation learning model,  in particular cases it suffers from the  \emph{uninformative representation} issue: the representation does not separate out the generative factors, and generator model relies on the weights information.

Following the direction suggested in \cite{alemi2017fixing,zhao2017infovae} we propose an information theoretic analysis of the VAE. Such description lead us to two observations: 
it is possible to learn both an informative representation and a generative model, and that is necessary to bound the network capacity in order to have a generator that does not relies on the weights and then more robust to noise \cite{JMLR:v19:17-646}. In light of this two observations we suggest to optimise the VAE according to the the Variational InfoMax (VIM) a variational objective, lower bound of the theoretical principlee: Capacity Constrained InfoMax (CCIM), ensuring to learn a maximally informative generator while maintaining bounded the network capacity.

The theoretical deductions, that inference and generative tasks are not orthogonal, and that an high capacity network although more informative, is prone to overfit, are confirmed by the computational experiments, where we compare the principal variants of the classic VAE \cite{alemi2017fixing}, \cite{higgins2017beta},  with two AutoEncoders, having different network capacity, optimising VIM. 

We conclude this section summarising the contribution of the paper in the following points: \begin{itemize}
    \item derivation of a variational lower bound for the maximal mutual information of a generative model belonging in a certain family, see \eqref{VIM};
    \item proposal of a new learning principle for unsupervised models: the Capacity-Constrained InfoMax, see \eqref{obj}, that allows both to learn a good representation while maintaining optimal generative performance;
    \item highlight the role of the latent entropy as a bound of the network capacity, see \eqref{obj};
    \item observe that a small capacity network is more robust to noise and then associated with a better inference model; see experiment section.
    
\end{itemize}

The work is divided as follows: in the second section we describe briefly the VAE and its variants; in the third and fourth sections we describe the variational infomax method and related work. We conclude the paper with the experimental results and the final observations.

\section{Background}

\subsection{Notation and preliminary definitions }
We use calligraphic letters (i.e. $\mathcal{X}$) for sets, capital letters (i.e. $X$) for random variables, and lower case letters (i.e. $x$) for their samples. With abuse of notation we denote both the probability and the corresponding density with the lower case letters (i.e. $p(x)$). 

\paragraph{KL divergence} Given two random distributions $p(x)$ and $q(x)$, the Kullback-Leibler (KL) divergence \begin{equation}
    D_{KL}(p(x)||q(x)) = \int \log \Big( \frac{p(y)}{q(y)}\Big)p(y)dy
\end{equation}
is an (intuitive) measure of the distance between the distributions $p$ and $q$. 

\paragraph{Mutual Information and Capacity} Given a channel $Z \to X$ with $X$ and $Z$ random variables, jointly distributed according to $p(x,z)$ and with marginals $p(x)$ and $p(z)$. The mutual information
 \begin{equation*}
    I(X,Z) = D_{KL}(p(x,z)||p(x)p(z)),
\end{equation*}
is a measure of the reduction of uncertainty in $X$ due to the knowledge of $Z$, and the capacity 
 \begin{equation*}
    C(X,Z) = \sup_{p(z) \in \mathcal{P}} I(X,Z)
\end{equation*}
is the maximal information that can be shared for a fixed generator $p(x|z)$. 

\subsection{Variational autoencoder}
From now on let us assume that the unknown distribution of the data $p(x)$ coincides with the empirical one $p_D(x)$, and that the distribution of the latent representation $p(z)$ is known. In this context the VAE is a model solving the following optimisation problem: find the generative model $p_\theta(x,z) \in \mathcal{P}_\theta$, specified by the parameters $\theta$ of the associated neural network, maximising the ELBO objective \begin{equation}\label{ELBO}
\begin{split}
    & ELBO_{\theta, \phi}  = \\
     &\quad \mathbb{E}_{p(x)}\mathbb{E}_{q(z|x)}[\log p(x|z)]
    - D_{KL}(q_\phi(z|x)||p(z))],
\end{split}
\end{equation}
a lower bound of the unfeasible-to-compute marginal likelihood $\mathbb{E}_{p(x)}[\log p_\theta(x)]$.
The ELBO objective is optimised by a regularised autoencoder, with encoder and decoder parametetrising, respectively, the inference and generative distributions, $q_\phi(z|x)$ and $p_\theta(x|z)$, with $\phi \in \Phi$, $\theta \in \Theta$  and regulariser defined by the \emph{rate} term $\mathbb{E}_{p(x)}[D_{KL}(q_\phi(z|x)||p(z))]$, an upper bound of the encoding information $I_\phi(Z,X) = \mathbb{E}_{p(x)}[D_{KL}(q_\phi(z|x)||q(z))]$.

\subsection{Uninformative representation issue}
The generator optimising the ELBO is the one such that its marginal $p_\theta(x)$ is minimising the divergence $D_{KL}(p_\theta(x)||p(x))$, a quantity that is independent from the inference distribution $q_\phi(\cdot|x)$ and the hidden representation $z$.  That means that optimising the ELBO objective does not guarantee an useful inference or generative model; indeed, in the case of really powerful generator model, the following catastrophic scenarios, are not rare :
\begin{itemize}
\item useless generative model: for any representation $z$ it is generated a sample from $p(x)$, since $p_\theta(x|z) = p_\theta(x)$, i.e. the information about the generated variable $X$ come from the weights $\theta$,
\item uninterruptible representation: in the latent space is impossible to identify the generative factors, since the learned representations are independent from the visible data, i.e. $q_\phi(z|x) = q_\phi(z)$.
\end{itemize}

Since both the scenarios are associated to a null information between $X$ and $Z$, respectively $I_\theta(X,Z)$ and $I_\phi(X;Z)$ and observing, by Data Processing Inequality \cite{cover2012elements},  that $I(g_\theta(Z), Z)\leq I(Z, X)$, 
in the next section we derive a variational objective, learning a maximal informative generator.

\section{The Model}
\subsection{The Variational InfoMax}
Assuming the distribution associated to the two random variable $O$ is known and $p(z) = p(o)$, the InfoMax objective is defined as: find the joint distribution $p_\theta(x,z) \in \mathcal{P}_\theta := \{p_\theta(x,z): \mathbb{E}_{p(z)}[p_\theta(x|z)] = p(x), \quad \mathbb{E}_{p(x)}[p_\theta(z|x)] = p(z) \}$ maximising the mutual information $I_\theta(X,Z) = D_{KL}(p_\theta(x,z)||p(x)p(z))$, i.e. find $\theta^* \in \Theta$ s.t. $I_{\theta^*} \geq I_\theta$ for any $\theta \in \Theta$.

Since the definition via KL divergence is computationally intractable, it is necessary to re-write the mutual information as \begin{equation}
    I_\theta(X,Z) = h_\theta(X) - h_\theta(X|Z),
\end{equation}
where $h_\theta(X) = -\mathbb{E}_{p_\theta(x)}[\log p_\theta(x)]$ is the entropy of $X$, and $h_\theta(X|Z) = -\mathbb{E}_{p_\theta(x,z)}[\log p_\theta(x|z)]$ is the conditional entropy $h_\theta(X|Z)$.
Since $p_\theta(x,z) \in \mathcal{P}_\theta$ the entropy $h_\theta(X) = h(X)$ is constant, and in order to maximise the mutual information it is sufficient to minimise the conditional entropy.

Excluding some special cases \cite{bell1997independent}, minimising the conditional entropy $ h_\theta(X|Z)$ is unfeasible, so it is necessary to consider an associated variational problem: for any $q_\phi(z|x)$ such that $q_\phi(z) = p(z)$,  learn the generative model $p_\theta(x|z)$ minimising the reconstruction accuracy term $h_{\theta,\phi}(X|Z) = \mathbb{E}_{p(x)}[\mathbb{E}_{q_\phi(z|x)}[\log(p_\theta (x|z))]]$.
Indeed, the cross-entropy
\begin{equation}\label{agk}
\begin{split}
    h_{\theta,\phi}(X|Z) & = h_\phi(X|Z) +  D_{KL}(q_\phi(z|x)||p_\theta(z|x))\\ &  \text{s.t } q_\phi(z) = p(z),
\end{split}
\end{equation}
is minimised when $q_\phi(z|x) = p_\theta(z|x) = p_{\theta^*}(z|x)$.

Unfortunately, the objective in \eqref{agk} is still unfeasible to compute, because it requires that $q_\phi(z) = p(z)$, but by the butterfly architecture of the autoencoder, $q_\phi(z)$ tends to be uniformly distributed on the space $\mathcal{Z}$. For this reason, we have to consider the following relaxed form:
\begin{equation}\label{VIM}
    VIM_{\theta, \phi} = h_{\theta,\phi}(X|Z) - \lambda D(q_\phi(z)||p(z)), \quad \lambda>0
\end{equation}
where it is introduced a term $D(q_\phi(z)||p(z))$ encouraging the empirical distribution $q_\phi(z)$ to be close, according to the metric $D$, to $p(z)$. 
In the following we assume $D = D_{KL}$, and in order to avoid any confusion the variational autoencoder trained maximising \eqref{VIM} will be dubbed VIMAE.

The derived objective is learning an maximally informative decoder, but by description \eqref{VIM} is not clear if the autoencoder learns an useful representation. To answer this question we have to consider the following equivalent description, \cite{zhao2017infovae}: 
\begin{equation}
\begin{split}\label{varVim}
     VIM_{\theta, \phi} = & -D_{KL}(p(x)||p_\theta(x)) - D_{KL}(q_\phi(z|x)||p_\theta(z|x))\\ & -(\lambda-1)D_{KL}(q_\phi(z)||p(z))+ I_{\phi}(X,Z).
\end{split}
\end{equation}
Thanks to the dual definition \eqref{varVim}, we see that the generator $p_\theta(x|z)$ is an actual generator since its marginal is close to the visible distribution (first term), and that the learned representation is maximally informative (fourth term), with maximal information bounded by the entropy $h_\theta(z)$ (third term), and finally that  the generative model does not relie on the weight information (second term), indeed if by contradiction $p_\theta(x|z) = p_\theta(x)$, we have a minimal encoding information, $I_\phi = 0.$

\subsection{Channel capacity}
In the ideal setting above, where the parameter families and the correct prior was known, we derived that the optimal solution is obtained by the generator having mutual information coincident with the network capacity 
\begin{equation}\label{capacity}
    C_{\theta}(X,Z) = \sup_{\theta, p(z) \in \mathcal{P}} I_{\theta}(X,Z).
\end{equation} 
Since, the decoding information is bounded by the inference distribution, see \eqref{varVim}, and the latter cannot grow more than the entropy $h(Z)$, optimal case that holds true in the no-noise channel case, we can assert that the VIM solution is the one optimising the following objective:
\begin{equation}\label{obj}
    \max_{\theta} I_\theta(X,Z)\quad \text{s.t.} \quad C_\theta(X,Z)  = h(Z).
\end{equation}
Let us observe that the capacity constrain is fundamental to ensure that the information about the generated variable $X$ came from the variable $Z$ and not from the weights $\theta$; indeed, the generative information $I_\theta(X;Z)$ can grow potentially up to $h(X)$, $p_\theta(x|z) = p_\theta(x)$, when the information about $Z$ is at most $h(Z)$. In the case where the generative model relies on the weight information we will say that the model is overfitting, \cite{JMLR:v19:17-646}.

By description above, we see that the choice of the, generally unknown, prior $p(z)$ plays a determinant role in the learning performance. Indeed, the relationship between capacity and network entropy suggests that a network with high entropy is more prone to overfit: the inference network is leaning unnecessary property about the data and the generative model relies on the weights information, $I(X, \theta)$.
In order to test the deduction that an high capacity network is more prone to overfit, in the experiments (see below) we consider the cases $Z$ is Normal (VIMAE-n) or Logistic (VIMAE-l) distributed. We choose to compare the popular Normal distribution with the Logistic one for two reasons:  the Logistic has less entropy than a Gaussian distribution and because it is a common assumption in natural science to suppose that the hidden factors of the visible data are logistically distributed \cite{hyvarinen2009natural}.

\section{Related work}
\paragraph{Autoencoder literature}
Autoencoder models are one of the most used family of neural networks to extract features in an unsupervised way \cite{bengio2013representation}, and their relationship with Information Theory is well-established from the first unregularised autoencoders \cite{baldi1989neural}.
The classical unregularised autoencoders, minimising the reconstruction loss $\mathbb{E}_{p(x)}[\mathbb{E}_{q_\phi(z|x)} [-\log p_\theta(x|z)]]$, are maximising an unbounded information, i.e. they are looking for a solution in the space $\tilde{\mathcal{P}}_\theta = \{p_\theta: p_\theta(x) = p(x)\}$. In general, a solution in this wide space is good only for reconstruction performance because $Z$ contains all the possible information that can be stored in the space $\mathcal{Z}$, and is not robust to input noise \cite{vincent2008extracting}; but, as observed in \cite{grover2018uncertainty}, if $q_\phi(z|x)\sim \mathcal{N}(\mu(x), \sigma(x))$ the model $p_\theta(x|z)$ is robust to noise and is a Gaussian generator. In this context the uninformative issue is avoided, but the price to pay is the impossibility to sample directly from a prior $p(z)$ that is not defined; indeed, the model described in \cite{grover2018uncertainty} requires running relatively expensive Markov Chain to obtain samples.

Many regularised models have been proposed, but the most well known is VAE, that minimises the expected code length of communicating $x$. As we observed in the previous sections, it is not guaranteed that the method finds a useful representation.
Such issue can be solved both controlling the information of the model, or considering a more flexible prior $p(z)$,\cite{rezende2015variational,kingma2016improved,dinh2016density}; the latter approach with free-inference model is the one obtaining best generative performance, but the inference model is difficult to interpret and to be used for different tasks from which was trained  \cite{achille2018life}, for this reason, in this manuscript we do not consider the latter approach.
\paragraph{VAE alternatives}
As observed above, a wrong choice of the prior $p(z)$ can be associated to a big encoding information and then to a not useful inference. For this reason in \cite{higgins2017beta} was proposed the $\beta$-VAE, a variational autoencoder optimising the following variant of the ELBO \begin{equation}\label{beta}
    \mathbb{E}_{q(z|x)}[-\log p_\theta(x|z)] - \beta \mathbb{E}_x [D_{KL}(q(z|x)||p(z))], \quad \beta>1.
\end{equation}
In this way it is controlled the mutual information $I_\phi(Z;X)$, obtaining a better inference.  Unfortunately in this way it is high the risk of an informative  generative model, having both pour generative quality and reconstruction accuracy.  Indeed, bounding the encoding information while maintaining a fixed high entropy prior, means that the reconstruction term is minimised relying on the weights information.
Moreover the choice of the parameter $\beta$ is tricky, since to an higher $\beta$ corresponds an higher probability to have a non-informative representation, $D_{KL}(q(z|x)||p(z))$. For this reason, in order to have both good inference and generative performance was suggested to optimise the following objective: \begin{equation}\label{Alemi}
        \mathbb{E}_{q(z|x)}[-\log p_\theta(x|z)] - \beta |C-\mathbb{E}_x [D_{KL}(q(z|x)||p(z))]|.
\end{equation}
In this way it is guaranteed to learn an inference information $I_\phi(X,Z)$ lower than $C$. Let us observe that the objective \eqref{Alemi} coincides with the VIM \eqref{VIM} in the case $C = h(Z)$, but differently from VIM, this approach has two main issues, one theoretic and the other computational. Theoretically, we solve only the uninformative issue, maintaining high the risk of a weight dependent generator, for the same reason of the $\beta$-VAE. Computationally, the learning principle \eqref{Alemi} is often intractable. 

Starting from different research point of views, respectively minimal cost generation and maximally informative inference, the objective in \eqref{VIM} was firstly derived in   \cite{zhao2017infovae} and\cite{tolstikhin2017wasserstein}. The main difference between our manuscript and the cited researches lies on the information theoretical analysis, and in particular on the description of the network capacity role and its relationship with the entropy of the latent prior. 
Finally we underline that is possible to consider in \eqref{VIM} distance measures different from the Kullback-Leibler divergence, for example in case we wish to consider a Jensen-Shannon divergence in \eqref{VIM} it is necessary to consider an adversarial network model, discriminating the true samples $z \sim p(z)$ from the fake sampled by $q_\phi(z)$ \cite{goodfellow2014generative}. In the latter case the obtained model is equivalent to the Adversarial AutoEncoder \cite{makhzani2015adversarial}.

\paragraph{Information theoretic literature} Information theory is strongly related with neural networks, and not only with autoencoders. Originally the InfoMax objective was applied to a self-organised system with a single hidden layer, \cite{bell1997independent,linsker1989application} where the bound in the capacity was given by the numbers of hidden neurons. More recently, the (naive) InfoMax has given way to a new information-theoretic principle: the Information-Bottleneck \cite{tishby2000information}. The idea of this principle is that a feed-forward neural network trained for task $T$ tends to learn a minimal sufficient representation of the data, maximising the following objective: \begin{equation}\label{IB}
    \max_Z I(Z,T) - \beta I(X,Z).
\end{equation} 
Although it was shown that in the general case this principle does not hold true \cite{saxe2018information}, the principle was used as a regularisation technique with success both in unsupervised \cite{alemi2017fixing,higgins2017beta} and supervised \cite{alemi2016deep} settings. 
We observe that the CCIM, \eqref{obj}, and IB, \eqref{IB}, coincide in the case of a deterministic encoder, where the encoding information is the entropy of $Z$.

\section{Experiments}
In this section we empirically evaluate both the generative and inference model learned optimised VIM \eqref{VIM}, we highlight the relationship between the network capacity and robust inference, and we compare the two VIMAE variants: VIMAE-n ($p(z)\sim \mathcal{N}(0,1)$) and VIMAE-l ($p(z)\sim Logistic(0,1)$), with the solution learned by ELBO \eqref{ELBO} and its principal variants \eqref{beta} and \eqref{Alemi}. 
In all the described experiments, the divergence $D_{KL}(q(z)||p(z))$ in \eqref{VIM} is approximated via the Maximum Mean Discrepancy \cite{zhao2017infovae} defined as:
\begin{equation}
    \text{MMD}(q(z), p(z)) = \sup_{f: \|f\|_{\mathcal{H}_k}\leq 1}\mathbb{E}_{p(z)}[f(Z)] - \mathbb{E}_{q(z)}[f(Z)]
\end{equation} 
where $\mathcal{H}_k$ is the Reproducing Kernel Hilbert Space associated to a positive definite kernel $k(\cdot, \cdot):\mathcal{Z}\times \mathcal{Z} \to \mathbb{R}_+$, and $f$ a map living in $\mathcal{H}_k$, i.e. $f: \mathcal{Z} \to \mathbb{R}$ such that $\langle f, k(x, \cdot)\rangle_{\mathcal{H}} = f(x)$. 

Moreover, by difficulties to compute the objective \eqref{Alemi}, as suggested in \cite{alemi2017fixing} we decided to optimise a $\beta$-VAE, denoted $\beta_A$-VAE, with $\beta<1$; in order to avoid any confusion, the original version proposed in \cite{higgins2017beta} with $\beta\gg1$ will be renamed $\beta_H$-VAE.

The experiments were performed with the same settings and autoencoder models used in \cite{tolstikhin2017wasserstein}, an architecture similar to the DCGAN \cite{radford2015unsupervised} with batch normalisation \cite{ioffe2015batch} (more details given in the Appendix). We consider four data-sets: MNIST, CIFAR10, and Omniglot three standard data-sets with ground-truth labels, to evaluate both the generative and inference models; and the CelebA \cite{liu2015deep}, a large entropic dataset consisting of roughly of 203k faces of $64\times 64$ resolution, in order to evalute the generative performance.

After considering many parameters for $\beta_H$, $\beta_A$ and $\lambda$, we choose, in accordance with what was suggested in \cite{tolstikhin2017wasserstein},  $\beta_H = \lambda = 10$, and $\beta_A = 0.2$ for MNIST and Omniglot and $ \beta = \lambda = 100$ and $\beta_A = 0.4$ for CelebA and CIFAR10 experiments. 

\subsection{Decoding information}

In this section we estimate the informativeness of the learnt generative model $p_\theta(x|z)$, evaluating both the generated sample quality and reconstruction accuracy of the associated model.

Given a representation variable $Z \sim p(Z)$, a model $p_\theta(\cdot|z)$ is said a good generator if the generated random variable $X_g \sim p_\theta(X)$, is close with respect a distance measure $D$, to the visible random variable $X_v \sim p(X)$. To evaluate the similarities between the generated sample and visible data, we consider two classic metrics: the Negative Log-Likelihood (NLL) for the grey-scale pictures, and the Frechet Inception Distance (FID), for the RGB datasets.  
The reason why we consider two different measures is twofold: firstly because FID, an estimation of the Frechet distance $\| X_g -X_v \|_2^2$ does not work well in the gray-scale setting, and secondly to highlight that a model with minimal NLL, or equivalently with minimal divergence $D_{KL}(p(x)||p_\theta(x))$, is not often the most informative or with sharper samples.

\begin{figure}
     \centering
     
     \begin{subfigure}[b]{0.3\textwidth}
         \centering
         \includegraphics[width=.45\textwidth]{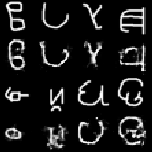}\hfill
         \includegraphics[width=.45\textwidth]{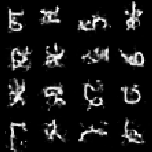}
         \caption*{VAE}
     \end{subfigure}
     
     \begin{subfigure}[b]{0.3\textwidth}
         \centering
         \includegraphics[width=.45\textwidth]{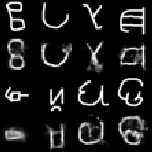}\hfill
         \includegraphics[width=.45\textwidth]{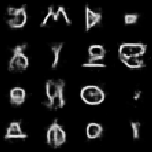}
         \caption*{$\beta_H$-VAE}
     \end{subfigure}
          \begin{subfigure}[b]{0.3\textwidth}
         \centering
         \includegraphics[width=.45\textwidth]{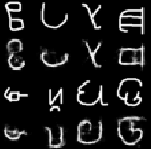}\hfill
         \includegraphics[width=.45\textwidth]{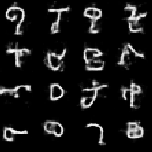}
         \caption*{$\beta_A$-VAE}
     \end{subfigure}
     \begin{subfigure}[b]{0.3\textwidth}
         \centering
         \includegraphics[width=.45\textwidth]{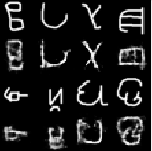}\hfill
         \includegraphics[width=.45\textwidth]{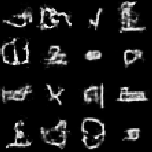}
        \caption*{VIMAE-n}

     \end{subfigure}
     \begin{subfigure}[b]{0.3\textwidth}
         \centering
         \includegraphics[width=.45\textwidth]{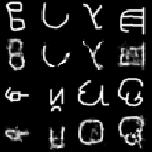}\hfill
         \includegraphics[width=.45\textwidth]{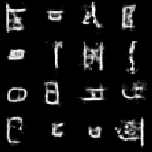}
         
         \caption*{VIMAE-l}
     \end{subfigure}

        \caption{Test reconstruction (left) and random generative samples (right) of the different methods with Omniglot. In test reconstructions, the odd rows are the original data. }
        \label{gen_rec_omn}

\end{figure}

\begin{figure}
     \centering
     
     \begin{subfigure}[b]{0.3\textwidth}
         \centering
         \includegraphics[width=.45\textwidth]{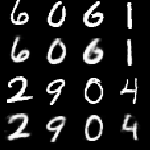}\hfill
         \includegraphics[width=.45\textwidth]{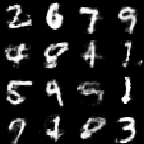}
         \caption*{VAE}
     \end{subfigure}
     
     \begin{subfigure}[b]{0.3\textwidth}
         \centering
         \includegraphics[width=.45\textwidth]{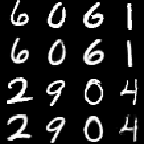}\hfill
         \includegraphics[width=.45\textwidth]{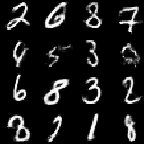}
         \caption*{$\beta_H$-VAE}
     \end{subfigure}
          \begin{subfigure}[b]{0.3\textwidth}
         \centering
         \includegraphics[width=.45\textwidth]{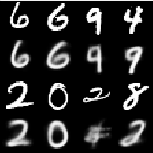}\hfill
         \includegraphics[width=.45\textwidth]{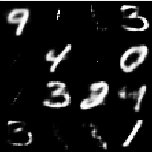}
         \caption*{$\beta_A$-VAE}
     \end{subfigure}
     \begin{subfigure}[b]{0.3\textwidth}
         \centering
         \includegraphics[width=.45\textwidth]{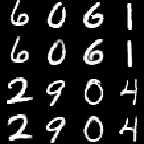}\hfill
         \includegraphics[width=.45\textwidth]{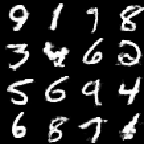}
        \caption*{VIMAE-n}

     \end{subfigure}
     \begin{subfigure}[b]{0.3\textwidth}
         \centering
         \includegraphics[width=.45\textwidth]{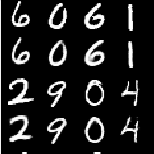}\hfill
         \includegraphics[width=.45\textwidth]{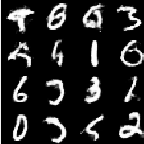}
         
         \caption*{VIMAE-l}
     \end{subfigure}

        \caption{Test reconstruction (left) and random generative samples (right) of the different methods with MNIST. In test reconstructions, the odd rows are the original data. }
        \label{gen_rec_mnist}

\end{figure}

Let us start considering the experiments on grey-scale dataset (MNIST and Omniglot), although the NLL associated to each model is similar, see table \ref{NLL}, we observe from figures \ref{gen_rec_omn} and \ref{gen_rec_mnist} that the quality of the generated samples differ. Indeed the samples generated by ELBO models are blur, that is because the information between the weights and the generated sample is high and then the generated data $\{ x_i\}_i$ are close to their average value $\bar{x} = \mathbb{E}[p_\theta(x)]$, this is consistent with the ELBO objective, indeed $D_{KL}(p(x)||p_\theta(\bar{x})) = 0$. This phenomenon appears particularly clear in the MNIST setting, where by the simplicity (small entropy) of the dataset, a model like $\beta_H$-VAE bounding the information between $X$ and $Z$ obtains optimal NLL performance.
\begin{table}
  \centering

\caption{NLL for generated samples (smaller is better)}

  \label{NLL}
  \begin{tabular}{lcccc}
    \toprule
    {}&     \multicolumn{2}{c}{NLL} &     \multicolumn{2}{c}{FID}   \\
    \cmidrule(r){2-3} \cmidrule(r){4-5}
    Method & MNIST & Omniglot & CIFAR10 & CelebA              \\
    \cmidrule(r){2-2} \cmidrule(r){3-3} \cmidrule(r){4-4} \cmidrule(r){5-5}

    VAE &  1158 & 1224 & 168 &  82\\
    $\beta_H$-VAE & \textbf{1113} & 1254 & 262 & -  \\
    $\beta_A$-VAE & 1123 & 1228 & 174  & 89 \\
    VIMAE-n   & 1169 & \textbf{1190} & \textbf{103}  &  56 \\
    VIMAE-l & 1171 &  1223 & 104  &  \textbf{55}    \\

    \bottomrule
  \end{tabular}
\end{table}

\begin{table}
\caption{Reconstruction accuracy, $\|\cdot \|_2$ over 100 samples, (smaller is better)}

  \label{RA}
  \centering
  \begin{tabular}{lcccc}
    \toprule
    {}&     \multicolumn{4}{c}{Reconstruction accuracy, $\|\cdot \|_2$}    \\
    \cmidrule(r){2-5} 
    Method & MNIST & Omniglot & CIFAR10 & CelebA              \\
    \cmidrule(r){2-2} \cmidrule(r){3-3} \cmidrule(r){4-4} \cmidrule(r){5-5}

    VAE &  0.51 & 0.75 & 8.29 & 17.29\\
    $\beta_H$-VAE & 0.62 & 0.75 & 9.8 & -\\
    $\beta_A$-VAE & 0.5 & 5.7 & 89 & 17.6\\
    VIMAE-n   & \textbf{0.47} & \textbf{0.75} & \textbf{4.74} & \textbf{16.65}  \\
    VIMAE-l & 0.48 &  0.76 & 4.85 & 16.74\\

    \bottomrule
  \end{tabular}
\end{table}

\begin{figure}
    \centering

 \begin{subfigure}[b]{0.4\textwidth}
         \centering
         \includegraphics[width=.45\textwidth]{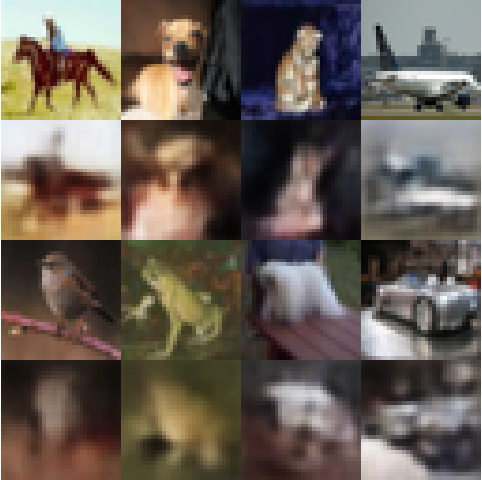}\hfill
         \includegraphics[width=.45\textwidth]{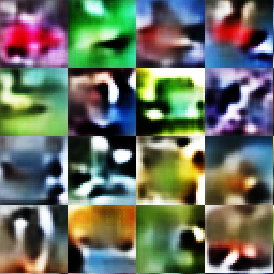}
         \caption*{ VAE}
     \end{subfigure}
     \begin{subfigure}[b]{0.4\textwidth}
         \centering
         \includegraphics[width=.45\textwidth]{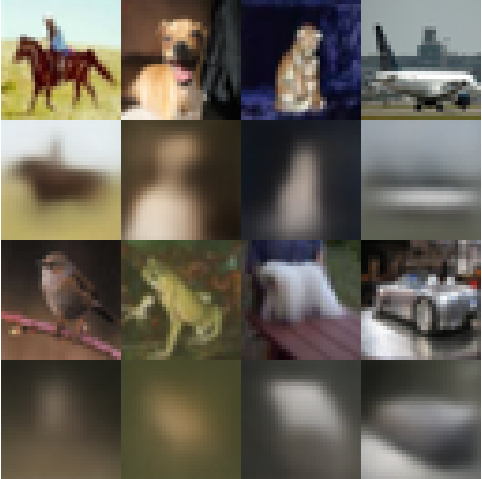}\hfill
         \includegraphics[width=.45\textwidth]{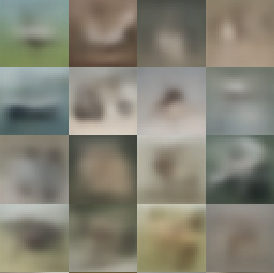}
         \caption*{$\beta_H$-VAE}
     \end{subfigure}
     \begin{subfigure}[b]{0.4\textwidth}
         \centering
         \includegraphics[width=.45\textwidth]{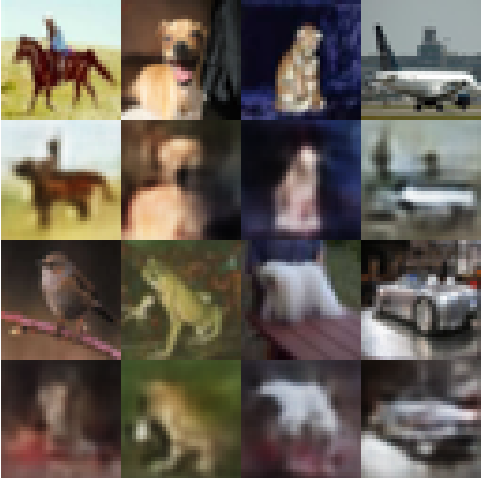}\hfill
         \includegraphics[width=.45\textwidth]{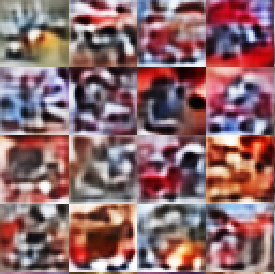}
         \caption*{$\beta_A$-VAE}
     \end{subfigure}
     \begin{subfigure}[b]{0.4\textwidth}
         \centering
         \includegraphics[width=.45\textwidth]{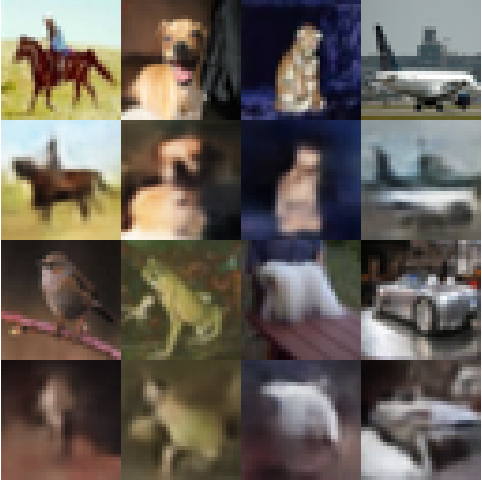}\hfill
         \includegraphics[width=.45\textwidth]{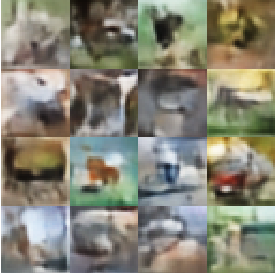}
         \caption*{VIMAE-n}
     \end{subfigure}
          \begin{subfigure}[b]{0.4\textwidth}
         \centering
         \includegraphics[width=.45\textwidth]{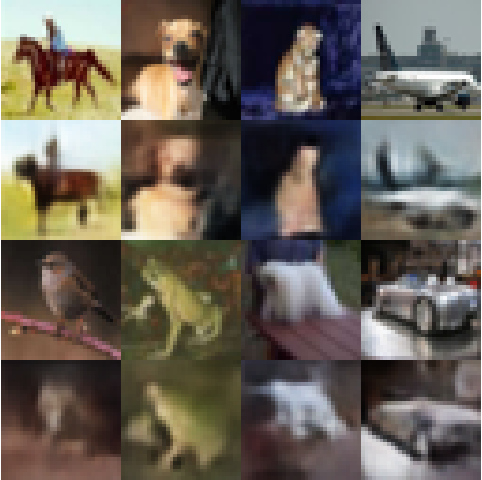}\hfill
         \includegraphics[width=.45\textwidth]{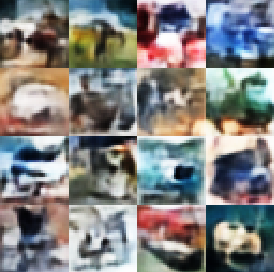}
         \caption*{VIMAE-l}
     \end{subfigure}
  \caption{Test reconstruction (left) and random generative samples (right) of the different methods with CIFAR10. In test reconstructions, the odd rows are the original data.}  \label{pic_cif}
\end{figure}

\begin{figure}
     \centering
     
     \begin{subfigure}[b]{0.4\textwidth}
         \centering
         \includegraphics[width=.45\textwidth]{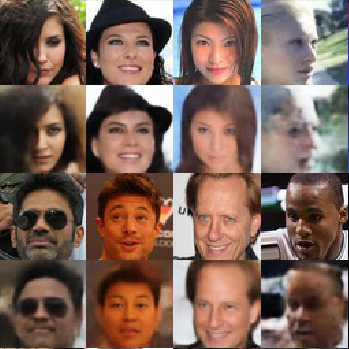}\hfill
         \includegraphics[width=.45\textwidth]{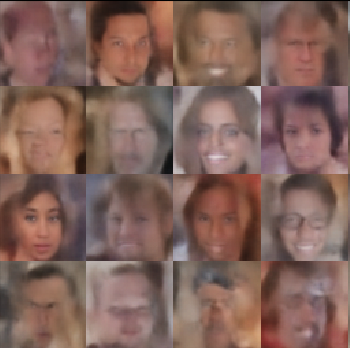}
         \caption*{VAE}
     \end{subfigure}
     
     \begin{subfigure}[b]{0.4\textwidth}
         \centering
         \includegraphics[width=.45\textwidth]{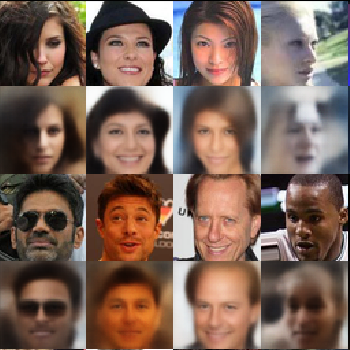}\hfill
         \includegraphics[width=.45\textwidth]{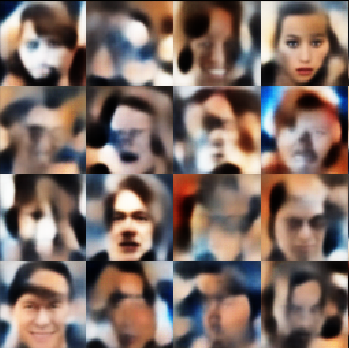}
         \caption*{$\beta_A$-VAE}
     \end{subfigure}

     \begin{subfigure}[b]{0.4\textwidth}
         \centering
         \includegraphics[width=.45\textwidth]{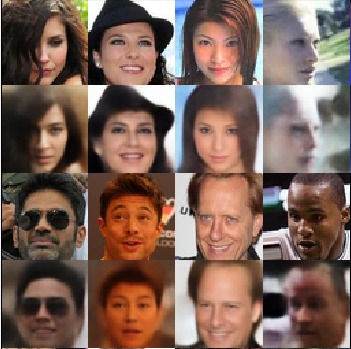}\hfill
         \includegraphics[width=.45\textwidth]{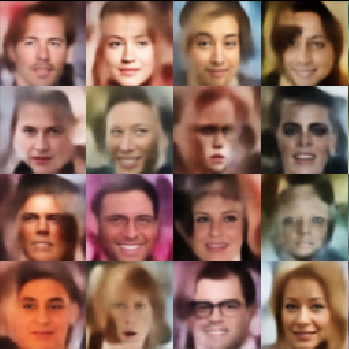}
        \caption*{VIMAE-n}

     \end{subfigure}
     \begin{subfigure}[b]{0.4\textwidth}
         \centering
         \includegraphics[width=.45\textwidth]{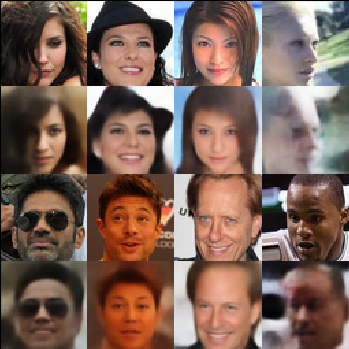}\hfill
         \includegraphics[width=.45\textwidth]{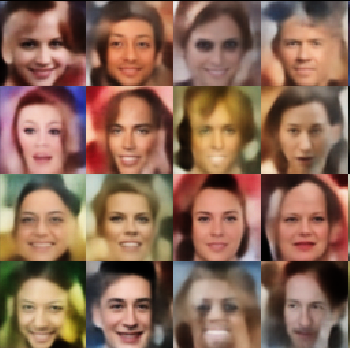}
         
         \caption*{VIMAE-l}
     \end{subfigure}

        \caption{Test reconstruction (left) and random generative samples (right) of the different methods with CelebA. In test reconstructions, the odd rows are the original data. }
        \label{gen_rec_celeba}

\end{figure}

More explanatory are the experiments in the challenging RGB setting. Where the VIM models have the best generative performance, as we can see from figures \ref{pic_cif}, \ref{gen_rec_celeba} and by the FID score in table \ref{NLL}. Moreover we observe that in this context $\beta_H$-VAE has poor results, for example in the CelebA we was not able to train a model with $\beta \gg 1$.

In order to confirm, that the VIMAE generators are the most informative, in table \ref{RA} we compare the reconstruction losses, a rude estimation of the decoding information, see \eqref{VIM}. According to the description made above we see, that apart from the Omniglot where all the models perform in a similar way, the VIMAE models have the best reconstruction performance in all the settings; in particular, in this task VIMAE-n performs better than VIMAE-l, in accordance to the idea that a small prior entropy is associated with a small capacity, and then less informative.

 Finally, we underline that the $\beta_H$-VAE, theoretically similar to the VIMAE behaves in different way, in particular it performs worse than classical VAE, this phenomenon is in agreement with the idea that a bigger capacity network tends naturally to overfit.

\subsection{Encoding information}
A good inference model  is  the one learning a representation where the generative factors are separate out (\emph{disentangled}) and robust to noise. In order to evaluate such properties, following the approach proposed in \cite{rifai2011higher}, we evaluate the accuracy of a supervised model directly trained on the feature space $\mathcal{Z}$. In particular, to evaluate the disentanglement property we consider the semi-supervised procedure used in \cite{zhao2017infovae}: we train the M1+TSVM \cite{kingma2014semi} on the feature data learnt by the autoencoder and use the classification accuracy over 1000 (100 for Omniglot) samples as an approximate metric to evaluate how much relevant are the learnt representation for a classification task. In order to evaluate the robustness of the learned features, we performed the same algorithm on the representation associated to corrupted data, i.e. $z \sim q(z| x + \nu)$, considering two types of noise: Gaussian and mask. In the Gaussian case, we add to each pixel a $\nu$ value sampled from $\mathcal{N}(0,\sigma^2)$ with $\sigma \in \{0.2, 0.3, 0.4\}$, and in the masking case a fraction $\nu$ of the elements is forced to be 0: each pixel is masked according to a Bernoulli distribution $\mathcal{B}(p), p \in \{0.2,0.5\}$. 
Higher classification performance suggests that the learned representation contains the relevant information and, in case of corrupted input data, that it is robust. In the Omniglot case by the challenge of the task (the test alphabet was never seen in the training) we consider a 5-character data-set, split into 300 ($60\times 5 $) for training and 100 for evaluation.

From the classification scores listed in tables \ref{smsup_cif}- \ref{smsup_omn},  we see that the ELBO-based model learnt good representations for clean data, but not when corrupted data is given as input.  This is particularly clear in the Bernoulli corruption case, that is a noise different from the one seen in the training. Particularly relevant is the behaviour of the two VIMAEs: they are comparable in the cases of clean data and small noise, but the one with big capacity, VIMAE-n, suffers in large noise setting, while the one with small capacity, VIMAE-l, is the most robust and in some challenging cases, see table  \ref{smsup_cif}, the noise helps to improve the model accuracy. Such a result is consistent with the idea that a small capacity network is learning the relevant factors of the input data, that are the only ones robust to the input noise.
\begin{table}
    \centering
      \caption{Semi-supervised classification CIFAR10.}

 \begin{tabular}{lccc}
    \toprule
    {}&     \multicolumn{2}{c}{accuracy (\%)}   \\
     \cmidrule(r){2-4}
         Method  & $\nu = 0$ & {$\mathcal{N}(0,0.3^2)$}  & $\mathcal{B}(0.2)$\\
    \cmidrule(r){1-4}
    VAE & {30} & 25 & 16\\
    $\beta_H$-VAE &  29 & 26 & 19\\
    $\beta_A$-VAE &  31 & 31 & 18\\
    VIMAE-n   & {29} & 28  & \textbf{23}  \\
    VIMAE-l & \textbf{32} & \textbf{34} & \textbf{23}  \\
    \bottomrule
  \end{tabular}

  \label{smsup_cif}

\end{table}

\begin{table}
      \caption{Semi-supervised classification, MNIST.}

  \centering
  \label{smsup}
  \begin{tabular}{lccccc}
    \toprule
    {}&     \multicolumn{5}{c}{accuracy (\%)}                   \\
    \cmidrule(r){2-6}
         Method  & $\nu = 0$ & \multicolumn{2}{c}{$\nu = \mathcal{N}(0,\sigma^2)$} & \multicolumn{2}{c}{$\nu = \mathcal{B}(p)$}  \\
    \cmidrule(r){1-1} \cmidrule(r){2-6}
          &  & $ 0.2$ &  $ 0.4$ &$ 0.2$ &$ 0.5$   \\
     \cmidrule(r){3-4}
      \cmidrule(r){5-6}
    VAE & 80 & 77 & 70 & 72 & 52\\
    $\beta_H$-VAE &  92 & 86 & 82& 91 & 84\\
     $\beta_A$-VAE &  \textbf{93} & 66 & 13& 85 & 65\\
    VIMAE-n   & \textbf{93} & \textbf{92} & 86 & \textbf{92}& 86   \\
    VIMAE-l & \textbf{93} & \textbf{92} & \textbf{88} & \textbf{92} & \textbf{87}\\
    \bottomrule
    \end{tabular}

\end{table}

\begin{table}
  \caption{Semi-supervised classification, Omniglot (random sampling: 20\%).}

    \centering
  \label{smsup_omn}
  \begin{tabular}{lccccc}
    \toprule
    & \multicolumn{5}{c}{accuracy (\%)}                   \\
    \cmidrule(r){1-1} \cmidrule(r){2-6}
          Method &  $\nu = 0$ & \multicolumn{2}{c}{$\nu = \mathcal{N}(0,\sigma^2)$} & \multicolumn{2}{c}{$\nu = \mathcal{B}(p)$}  \\
    \cmidrule(r){2-6}
           & {} & $ 0.2$ &  $ 0.4$ &$ 0.2$ &$ 0.5$   \\
     \cmidrule(r){3-4}
      \cmidrule(r){5-6}
     VAE & 22 & 22 & 17 & 22 & 16\\
      $\beta_H$-VAE & 21 & 21 & 22& 19 & 17\\
      $\beta_A$-VAE & 22 & 22 & 21& 21 & 24\\
     VIMAE-n & {22} & \textbf{23} & \textbf{24} & 22 & \textbf{22}  \\
     VIMAE-l & \textbf{24} & \textbf{23} & {20} & \textbf{23} & \textbf{22}       \\
    \bottomrule
    \end{tabular}

\end{table}
\section{Conclusion}

We observed, via an information theoretic description of VAE, that it is possible to learn a good generative model while maintaining a meaningful hidden representation, and that goal can be reached by optimising the CCIM, an objective that separates out the two properties of a network: the generative information and its capacity. We underlined the relationship between robustness and network capacity and how that one can be defined by the prior $p(z)$

The definition of the network capacity, and its strictly relationship with the choice of the latent prior, suggests that the VIMAE could be used in tasks where it is necessary to modify the network capacity continually. For example, in the Life-Long learning case where the choice of the network capacity is fundamental in order to avoid the catastrophic forgetting issue \cite{achille2018life}. 
In the light of the good performance of the CCIM objective, and its relationship with the Information Bottleneck, future work include the generalisation of the CCIM to the supervised case where the Information Bottleneck, is considered the best option \cite{alemi2016deep}. 
\bibliography{iclr2019_conference}

\begin{thebibliography}{10}
\providecommand{\url}[1]{#1}
\csname url@samestyle\endcsname
\providecommand{\newblock}{\relax}
\providecommand{\bibinfo}[2]{#2}
\providecommand{\BIBentrySTDinterwordspacing}{\spaceskip=0pt\relax}
\providecommand{\BIBentryALTinterwordstretchfactor}{4}
\providecommand{\BIBentryALTinterwordspacing}{\spaceskip=\fontdimen2\font plus
\BIBentryALTinterwordstretchfactor\fontdimen3\font minus
  \fontdimen4\font\relax}
\providecommand{\BIBforeignlanguage}[2]{{%
\expandafter\ifx\csname l@#1\endcsname\relax
\typeout{** WARNING: IEEEtran.bst: No hyphenation pattern has been}%
\typeout{** loaded for the language `#1'. Using the pattern for}%
\typeout{** the default language instead.}%
\else
\language=\csname l@#1\endcsname
\fi
#2}}
\providecommand{\BIBdecl}{\relax}
\BIBdecl

\bibitem{rifai2011higher}
S.~Rifai, G.~Mesnil, P.~Vincent, X.~Muller, Y.~Bengio, Y.~Dauphin, and
  X.~Glorot, ``Higher order contractive auto-encoder,'' in \emph{Joint European
  Conference on Machine Learning and Knowledge Discovery in Databases}.\hskip
  1em plus 0.5em minus 0.4em\relax Springer, 2011, pp. 645--660.

\bibitem{lipton2018mythos}
Z.~C. Lipton, ``The mythos of model interpretability,'' \emph{Queue}, vol.~16,
  no.~3, pp. 31--57, 2018.

\bibitem{dinh2016density}
L.~Dinh, J.~Sohl-Dickstein, and S.~Bengio, ``Density estimation using real
  nvp,'' \emph{arXiv preprint arXiv:1605.08803}, 2016.

\bibitem{hinton2006fast}
G.~E. Hinton, S.~Osindero, and Y.-W. Teh, ``A fast learning algorithm for deep
  belief nets,'' \emph{Neural computation}, vol.~18, no.~7, pp. 1527--1554,
  2006.

\bibitem{maddison2017filtering}
C.~J. Maddison, J.~Lawson, G.~Tucker, N.~Heess, M.~Norouzi, A.~Mnih, A.~Doucet,
  and Y.~Teh, ``Filtering variational objectives,'' in \emph{Advances in Neural
  Information Processing Systems}, 2017, pp. 6573--6583.

\bibitem{radford2015unsupervised}
A.~Radford, L.~Metz, and S.~Chintala, ``Unsupervised representation learning
  with deep convolutional generative adversarial networks,'' \emph{arXiv
  preprint arXiv:1511.06434}, 2015.

\bibitem{kingma2013auto}
D.~P. Kingma and M.~Welling, ``Auto-encoding variational bayes,'' \emph{arXiv
  preprint arXiv:1312.6114}, 2013.

\bibitem{rezende2014stochastic}
D.~J. Rezende, S.~Mohamed, and D.~Wierstra, ``Stochastic backpropagation and
  approximate inference in deep generative models,'' \emph{arXiv preprint
  arXiv:1401.4082}, 2014.

\bibitem{goodfellow2014generative}
I.~Goodfellow, J.~Pouget-Abadie, M.~Mirza, B.~Xu, D.~Warde-Farley, S.~Ozair,
  A.~Courville, and Y.~Bengio, ``Generative adversarial nets,'' in
  \emph{Advances in neural information processing systems}, 2014, pp.
  2672--2680.

\bibitem{achille2018life}
A.~Achille, T.~Eccles, L.~Matthey, C.~Burgess, N.~Watters, A.~Lerchner, and
  I.~Higgins, ``Life-long disentangled representation learning with
  cross-domain latent homologies,'' in \emph{Advances in Neural Information
  Processing Systems}, 2018, pp. 9873--9883.

\bibitem{ramapuram2017lifelong}
J.~Ramapuram, M.~Gregorova, and A.~Kalousis, ``Lifelong generative modeling,''
  \emph{arXiv preprint arXiv:1705.09847}, 2017.

\bibitem{alemi2017fixing}
A.~A. Alemi, B.~Poole, I.~Fischer, J.~V. Dillon, R.~A. Saurous, and K.~Murphy,
  ``Fixing a broken elbo,'' \emph{arXiv preprint arXiv:1711.00464}, 2017.

\bibitem{zhao2017infovae}
S.~Zhao, J.~Song, and S.~Ermon, ``Infovae: Information maximizing variational
  autoencoders,'' \emph{arXiv preprint arXiv:1706.02262}, 2017.

\bibitem{JMLR:v19:17-646}
\BIBentryALTinterwordspacing
A.~Achille and S.~Soatto, ``Emergence of invariance and disentanglement in deep
  representations,'' \emph{Journal of Machine Learning Research}, vol.~19,
  no.~50, pp. 1--34, 2018. [Online]. Available:
  \url{http://jmlr.org/papers/v19/17-646.html}
\BIBentrySTDinterwordspacing

\bibitem{higgins2017beta}
I.~Higgins, L.~Matthey, A.~Pal, C.~Burgess, X.~Glorot, M.~Botvinick,
  S.~Mohamed, and A.~Lerchner, ``beta-vae: Learning basic visual concepts with
  a constrained variational framework,'' in \emph{International Conference on
  Learning Representations}, vol.~3, 2017.

\bibitem{cover2012elements}
T.~M. Cover and J.~A. Thomas, \emph{Elements of information theory}.\hskip 1em
  plus 0.5em minus 0.4em\relax John Wiley \& Sons, 2012.

\bibitem{bell1997independent}
A.~J. Bell and T.~J. Sejnowski, ``The “independent components” of natural
  scenes are edge filters,'' \emph{Vision research}, vol.~37, no.~23, pp.
  3327--3338, 1997.

\bibitem{hyvarinen2009natural}
A.~Hyv{\"a}rinen, J.~Hurri, and P.~O. Hoyer, \emph{Natural image statistics: A
  probabilistic approach to early computational vision.}\hskip 1em plus 0.5em
  minus 0.4em\relax Springer Science \& Business Media, 2009, vol.~39.

\bibitem{bengio2013representation}
Y.~Bengio, A.~Courville, and P.~Vincent, ``Representation learning: A review
  and new perspectives,'' \emph{IEEE transactions on pattern analysis and
  machine intelligence}, vol.~35, no.~8, pp. 1798--1828, 2013.

\bibitem{baldi1989neural}
P.~Baldi and K.~Hornik, ``Neural networks and principal component analysis:
  Learning from examples without local minima,'' \emph{Neural networks},
  vol.~2, no.~1, pp. 53--58, 1989.

\bibitem{vincent2008extracting}
P.~Vincent, H.~Larochelle, Y.~Bengio, and P.-A. Manzagol, ``Extracting and
  composing robust features with denoising autoencoders,'' in \emph{Proceedings
  of the 25th international conference on Machine learning}.\hskip 1em plus
  0.5em minus 0.4em\relax ACM, 2008, pp. 1096--1103.

\bibitem{grover2018uncertainty}
A.~Grover and S.~Ermon, ``Uncertainty autoencoders: Learning compressed
  representations via variational information maximization,'' \emph{arXiv
  preprint arXiv:1812.10539}, 2018.

\bibitem{rezende2015variational}
D.~J. Rezende and S.~Mohamed, ``Variational inference with normalizing flows,''
  \emph{arXiv preprint arXiv:1505.05770}, 2015.

\bibitem{kingma2016improved}
D.~P. Kingma, T.~Salimans, R.~Jozefowicz, X.~Chen, I.~Sutskever, and
  M.~Welling, ``Improved variational inference with inverse autoregressive
  flow,'' in \emph{Advances in neural information processing systems}, 2016,
  pp. 4743--4751.

\bibitem{tolstikhin2017wasserstein}
I.~Tolstikhin, O.~Bousquet, S.~Gelly, and B.~Schoelkopf, ``Wasserstein
  auto-encoders,'' \emph{arXiv preprint arXiv:1711.01558}, 2017.

\bibitem{makhzani2015adversarial}
A.~Makhzani, J.~Shlens, N.~Jaitly, I.~Goodfellow, and B.~Frey, ``Adversarial
  autoencoders,'' \emph{arXiv preprint arXiv:1511.05644}, 2015.

\bibitem{linsker1989application}
R.~Linsker, ``An application of the principle of maximum information
  preservation to linear systems,'' in \emph{Advances in neural information
  processing systems}, 1989, pp. 186--194.

\bibitem{tishby2000information}
N.~Tishby, F.~C. Pereira, and W.~Bialek, ``The information bottleneck method,''
  \emph{arXiv preprint physics/0004057}, 2000.

\bibitem{saxe2018information}
A.~M. Saxe, Y.~Bansal, J.~Dapello, M.~Advani, A.~Kolchinsky, B.~D. Tracey, and
  D.~D. Cox, ``On the information bottleneck theory of deep learning,'' 2018.

\bibitem{alemi2016deep}
A.~A. Alemi, I.~Fischer, J.~V. Dillon, and K.~Murphy, ``Deep variational
  information bottleneck,'' \emph{arXiv preprint arXiv:1612.00410}, 2016.

\bibitem{ioffe2015batch}
S.~Ioffe and C.~Szegedy, ``Batch normalization: Accelerating deep network
  training by reducing internal covariate shift,'' \emph{arXiv preprint
  arXiv:1502.03167}, 2015.

\bibitem{liu2015deep}
Z.~Liu, P.~Luo, X.~Wang, and X.~Tang, ``Deep learning face attributes in the
  wild,'' in \emph{Proceedings of the IEEE international conference on computer
  vision}, 2015, pp. 3730--3738.

\bibitem{kingma2014semi}
D.~P. Kingma, S.~Mohamed, D.~J. Rezende, and M.~Welling, ``Semi-supervised
  learning with deep generative models,'' in \emph{Advances in neural
  information processing systems}, 2014, pp. 3581--3589.

\end{thebibliography}
\bibliographystyle{iclr2019_conference}
\end{document}